\documentclass[conference]{IEEEtran}
\usepackage{times}

\usepackage[numbers]{natbib}
\usepackage{multicol}
\usepackage[bookmarks=true]{hyperref}

\usepackage{graphics} 
\usepackage{graphicx}
\usepackage{subfigure}

\def\statespace {{\cal S}}
\def\mdp {{\cal M}}
\def\actionspace {{\cal A}}
\def\probs {{\cal P}}
\def\transitionmodel {{\cal T}}

\begin{document}

\title{Robot Perception enables Complex\\ Navigation Behavior via Self-Supervised Learning}


\author{\authorblockN{Marvin Chanc\'an and Michael Milford}
\authorblockA{Electrical Engineering and Robotics, QUT Centre for Robotics\\
Queensland University of Technology, Australia\\
Email: mchancanl@uni.pe
}}


%

\maketitle

\begin{abstract}
Learning visuomotor control policies in robotic systems is a fundamental problem when aiming for long-term behavioral autonomy. Recent supervised-learning-based vision and motion perception systems, however, are often separately built with limited capabilities, while being restricted to few behavioral skills such as passive visual odometry (VO) or mobile robot visual localization. Here we propose an approach to unify those successful robot perception systems for active target-driven navigation tasks via reinforcement learning (RL). Our method temporally incorporates compact \textit{motion and visual perception} data---directly obtained using self-supervision from a single image sequence---to enable complex goal-oriented navigation skills. We demonstrate our approach on two real-world driving dataset, KITTI and Oxford RobotCar, using the new interactive CityLearn framework. The results show that our method can accurately generalize to extreme environmental changes such as day to night cycles with up to an 80\% success rate, compared to 30\% for a vision-only navigation systems.
\end{abstract}

\IEEEpeerreviewmaketitle

\section{Introduction}
Recent advances in self-supervised learning have show promising results in a range of visuomotor tasks including robotic manipulation \cite{8014803,8593986, deng2019selfsupervised, IEEEexample:lee2020making, IEEEexample:pmlr-v87-ebert18a, IEEEexample:Nair2020Hierarchical} using deep reinforcement learning (RL), both in simulation and on real hardware. For mobile robots, these self-supervised learning techniques are now being explored and have already show to achieve comparable results to classical robot perception pipelines for passive visual odometry (VO) \cite{IEEEexample:zhan2019visual, IEEEexample:2020_Wagstaff_Self-Supervised}, visual localization or place recognition (VPR) \cite{IEEEexample:ge2020self}, and also active outdoor navigation tasks \cite{IEEEexample:kahn2020badgr} in real environments. Nevertheless, end-to-end
learning of visuomotor policies for long-term, all-weather
autonomous navigation tasks using self-supervision remains unexplored.


Large-scale outdoor navigation is a key component for enabling the deployment of mobile robots and autonomous vehicles in the real world. Recent RL-based navigation systems for real environments rely on GPS-based ground-truth data for labeling raw sensory images. They then reduce the problem of navigation to vision-only methods \cite{IEEEexample:NIPS2018_7509} or extend it with language-based sensory inputs. These approaches: 1) are generally hard to train---due to their weakly-related input sensor modalities, 2) rely on the precision of GPS data---which may not be reliable across month-spaced traversals of the same route, and 3) require a large amount of experience with the environment in terms of RL training episodes---which might be impractical for real robots. Moreover, their generalization capabilities to different environmental changes such as lighting or weather transitions are not explored. 


\begin{figure}[!t]
   \centering
   \includegraphics[width=\columnwidth]{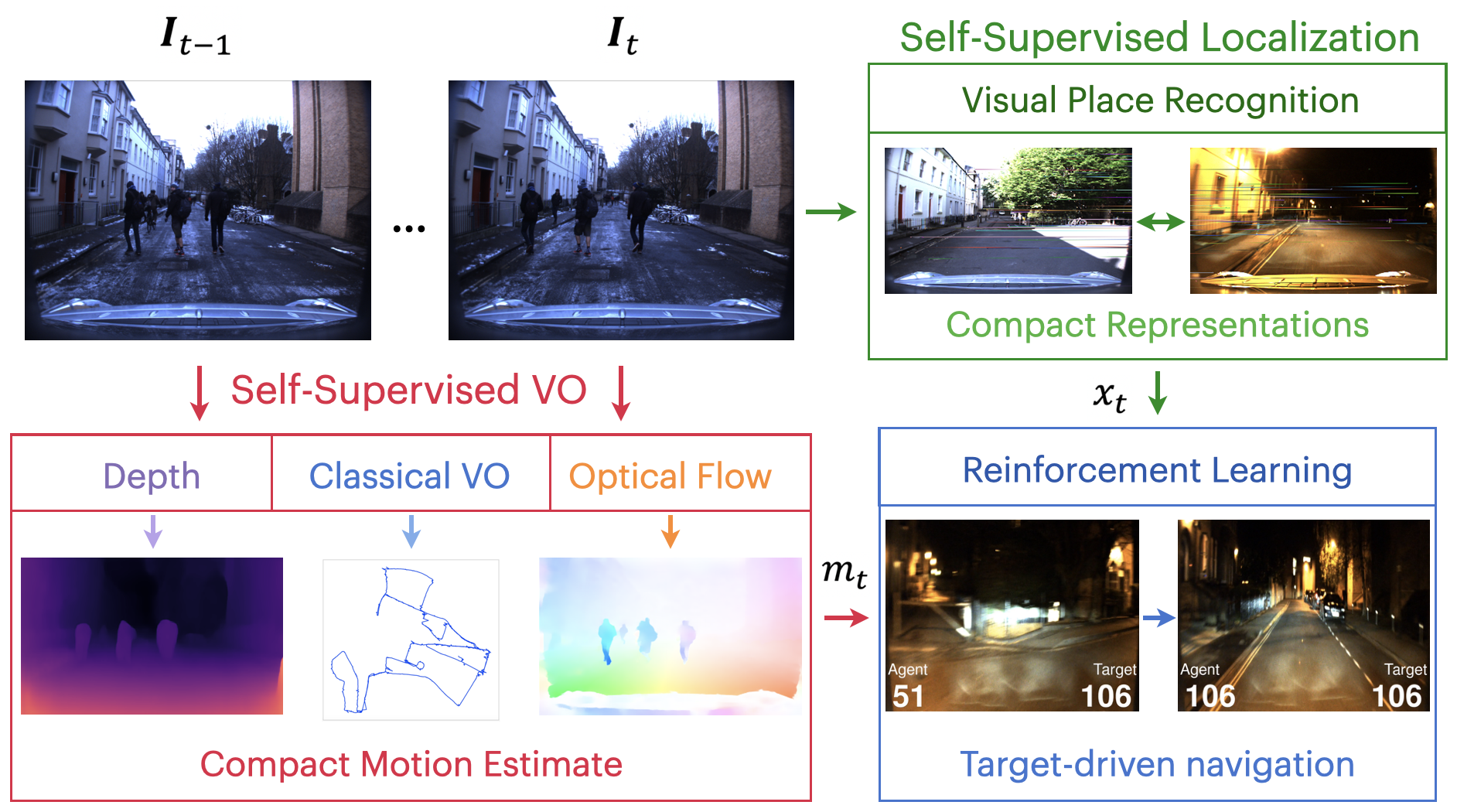}
   \vspace{-5mm}
   \caption{Overview of our unified robot learning framework for navigation tasks. Given a single traversal of a car ride ($\mathbf{I_t}$), we use self-supervised learning to obtain optimized VO data ($\mathbf{m_t}$) and visual representations ($\mathbf{x_t}$). We then temporally combine these compact visuomotor signals to learn control policies for goal-driven navigation skills via RL. Our method can accurately generalize to extreme environmental changes such as day to night transitions.}
   \label{approach}
   \vspace{-6mm}
\end{figure}

In this paper, instead of relying on supervised learning methods for capturing motion and visual representations, we investigate how to leverage recent self-supervised learning approaches for enabling efficient and robust long-term robot navigation skills. Our key contributions are:
\begin{itemize}
    \item An approach to temporally integrate motion states (classical VO self-optimized with optical flow and depth prediction) with visual observations (self-enhanced with image-to-region similarities) via RL for large-scale, all-weather navigation tasks (see Fig. \ref{approach}), and
    \item Experimental trade-off between the RL navigation success rate and the motion estimation precision, providing key insights to decide which ego-motion sensor would be appropriate for a particular application.
\end{itemize}

We demonstrate the effectiveness and advantages of our method on two large, real driving datasets for goal-oriented navigation tasks, compared to motion-only and vision-only navigation systems. Furthermore, we report experimental results where our approach is capable of generalizing to extreme environmental transitions such as day to night cycles with high navigation success rate, where vision-only navigation systems typically fail.

\section{Problem Formulation}
\label{sec:formulation}

We formulate the goal-driven navigation task as a Markov decision process $\mdp$: at any given discrete state $\mathbf{s}_t \in \statespace$ at time $t$, the robot executes a discrete action $\mathbf{a}_t \in \actionspace$ following the policy $\pi_\theta: \statespace \to \probs(\actionspace)$, then transitions to a new state $\mathbf{s}_{t+1}$ receiving a corresponding reward $r$. We train our policy to find an optimal $\theta^*$ that maximizes the objective function given by $\mathbf{E}_{\tau \sim \pi_\theta(\tau)} \sum_{t=1}^T \gamma r(\tau)$, with a transition operator $\transitionmodel: \statespace \times \actionspace \to \statespace$ and a $\gamma$-discounted reward function over a finite-horizon $T$.

In this work, following the main ideas proposed in \cite{chancan2020mvp} and \cite{IEEEexample:chancan2020hybrid}, we investigate how to temporally incorporate in $\statespace$ compact motion states, $\mathbf{m}_t$, with equally compact visual observations, $\mathbf{x}_t$, both obtained via self-supervised learning from a single monocular image sequence, $\mathbf{I}_t$, using the state-of-the-art RL algorithm PPO \cite{ppo} (see Fig. \ref{approach}).

\section{Approach}
\label{sec:approach}

Our objective is to train an RL agent to perform goal-driven navigation tasks across a range of real-world environmental conditions, especially where noise or poor GPS data typically limit the capabilities of supervised learning approaches. We therefore developed a combined motion-and-vision-based perception method that can be trained using self-supervision. Our approach operates by temporally associating local motion states, obtained from VO-based techniques, with visual observations to efficiently train our navigation policy network, Fig. \ref{approach}. This enables our policy to learn from both motion and visual information in a self-supervised manner, while training using an RL framework, to being robust to environmental visual changes and also poor GPS data.

\subsection{Self-Supervised Single-Frame Visual Localization}

In image-based localization, weak GPS- or geo-tagged labels can be problematic when training visual place recognition (VPR) systems using supervised learning. To overcome these challenges, successful VPR systems such as NetVLAD \cite{Arandjelovic_2016_CVPR} have achieved state-of-the-art results via weakly-supervised learning, with a range of recent developments \cite{8099829, Liu_2019_ICCV}. More recently, however, a self-supervised fine-grained region similarities (SFRS) system, especially designed for dealing with noisy pairwise image-label, has outperformed these VPR pipelines \cite{IEEEexample:ge2020self}. In this work, we attempt to merge the desirable properties of SFRS into our RL-based navigation system for leveraging image-to-region similarities when GPS labels are poor or not available for large-scale image perception.

\subsection{Self-Supervised Monocular Visual Odometry}

In robot navigation research, visual odometry (VO) and SLAM techniques are also typically used for performing visual-based localization; providing key complementary information of the environment along with GPS, IMU or LiDAR sensors. While SLAM extends VO, along with loop closing and global map optimization, for building a geometrically consistent map of the environment, VO continues to be a fundamental component for proving ego-motion estimate data for mobile robots. With the rapid progress of deep learning techniques in computer vision, roboticists have been attracted to incorporate these learning capabilities for VO over the past 4 years \cite{zhou2017unsupervised, 8461251}. Only recently, however, the use of more advanced self-supervised learning techniques have enabled to outperform those purely geometry-based or deep-learning-based VO systems \cite{IEEEexample:zhan2019visual, IEEEexample:2020_Wagstaff_Self-Supervised}. Here we incorporate a self-supervised deep pose corrections method (SS-DPC-Net) \cite{IEEEexample:2020_Wagstaff_Self-Supervised}, which combine depth estimation, optical flow, and classical VO in a hybrid manner, for robust VO into our RL-based system, providing compact and optimized ego-motion estimate data.

\begin{figure}[!t]
\centering
\subfigure{\includegraphics[width=0.32\columnwidth, height=14mm]{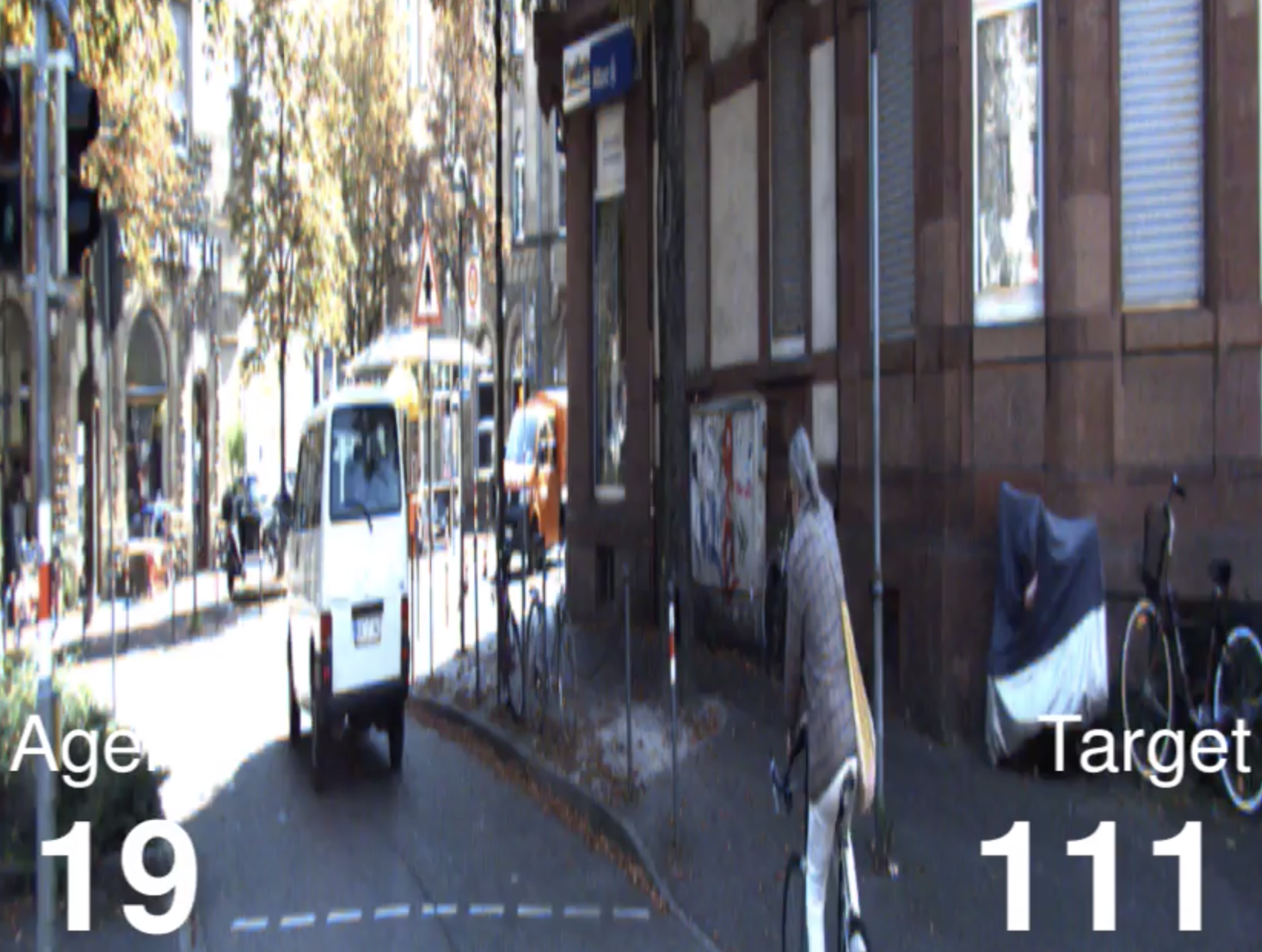}}
\subfigure{\includegraphics[width=0.32\columnwidth, height=14mm]{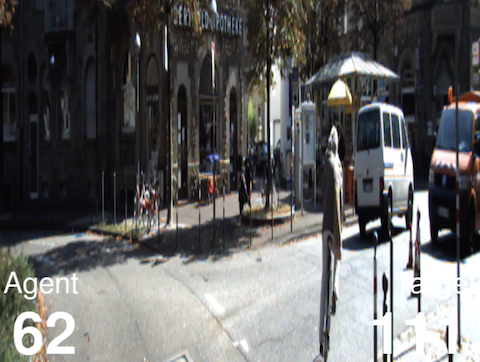}}
\subfigure{\includegraphics[width=0.32\columnwidth, height=14mm]{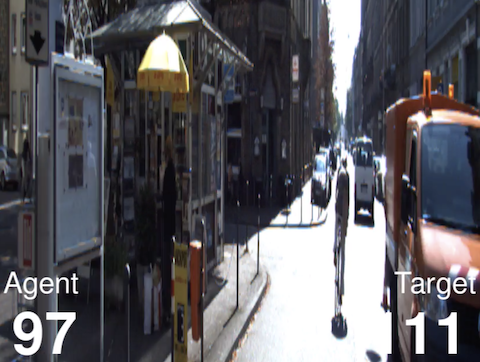}}
\vspace{-2mm}
\caption{\textbf{Deployment results} on the KITTI dataset. The agent navigates from left to right towards the goal destination.}
\label{kitti-deploy}
\vspace{-2mm}
\end{figure}

\begin{figure}[!t]
\centering
\subfigure{\includegraphics[width=0.32\columnwidth, height=14mm]{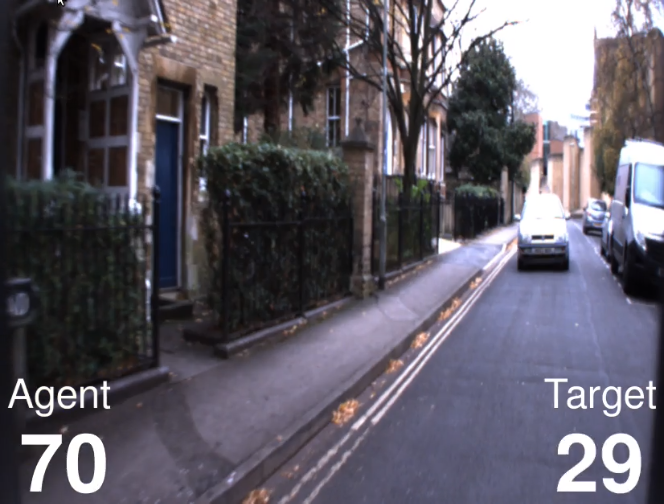}}
\subfigure{\includegraphics[width=0.32\columnwidth, height=14mm]{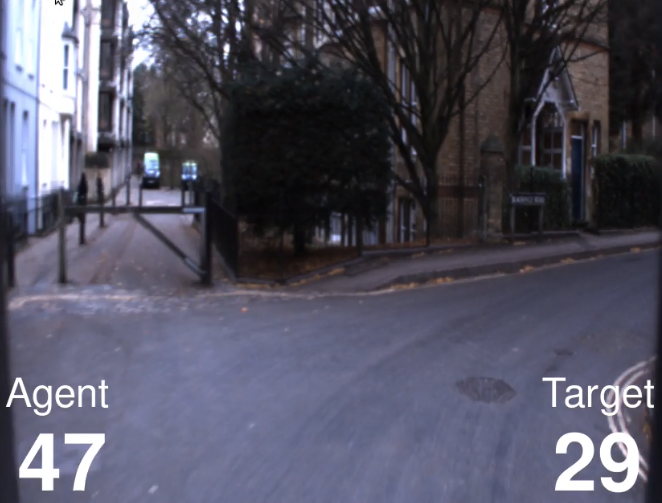}}
\subfigure{\includegraphics[width=0.32\columnwidth, height=14mm]{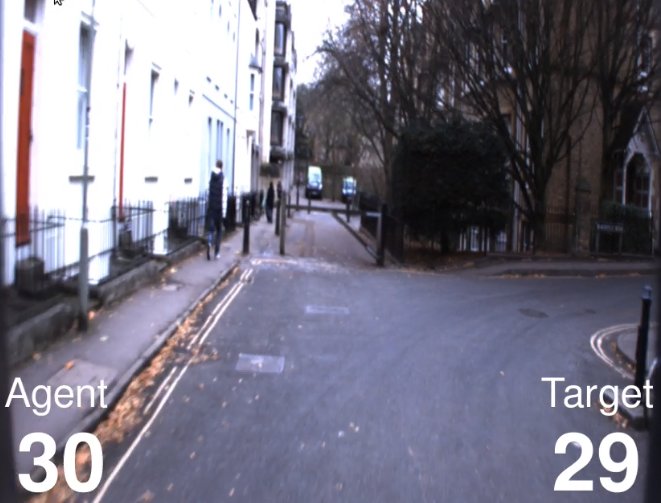}}
\subfigure{\includegraphics[width=0.32\columnwidth, height=14mm]{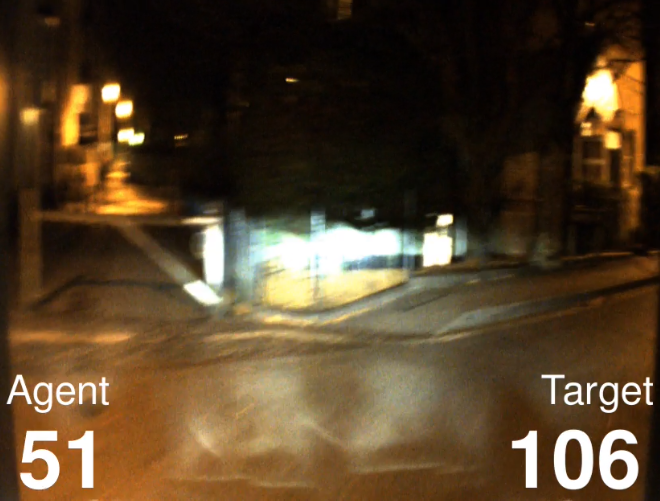}}
\subfigure{\includegraphics[width=0.32\columnwidth, height=14mm]{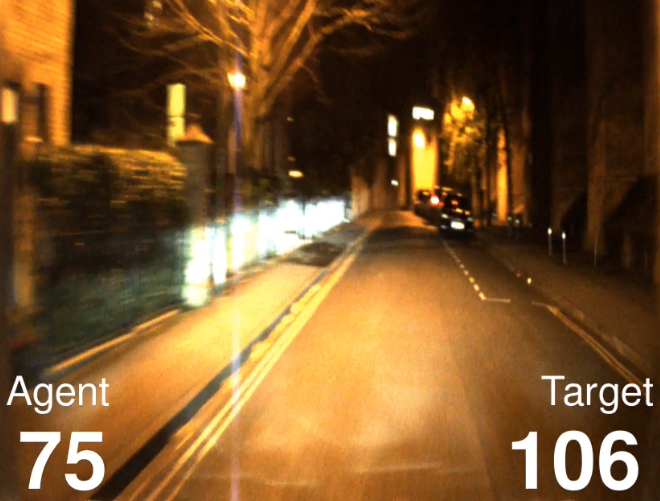}}
\subfigure{\includegraphics[width=0.32\columnwidth, height=14mm]{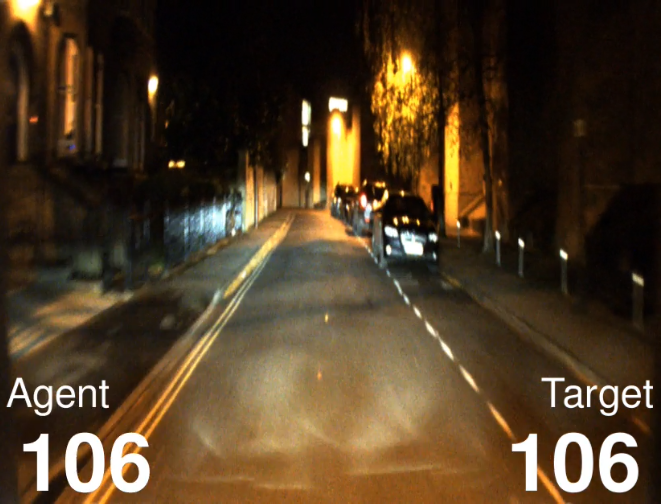}}
\vspace{-2mm}
\caption{\textbf{Deployment results} on the Oxford RobotCar dataset. The agent navigates from left to right towards the goal destination on the traversal it was trained (top), and generalize well at night (bottom).}
\label{ox-deploy}
\vspace{-6mm}
\end{figure}

\subsection{Reinforcement learning-based navigation}

\textbf{Goal-driven navigation}: We merge both motion states, $\mathbf{m}_t$, and visual observations, $\mathbf{x}_t$, obtained via self-supervision from raw image sequences, for learning to navigate through actions, $\mathbf{a}_t$, towards a required goal destination, $\mathbf{g}_t$, via RL \cite{ppo}.

\textbf{Architecture}: Our policy network is inspired by \cite{IEEEexample:NIPS2018_7509}, which includes a single \textit{linear} layer with $512$ units to encode $\mathbf{m}_t$ and $\mathbf{x}_t$. Then, using a single recurrent layer long short-term memory (LSTM) with $256$ units, current states and observations are combined with the agent's previous actions, $\mathbf{a}_{t-1}$. The updated agent's actions, $\mathbf{a}_{t}$, are then used to estimate both the new actions and the value function $V$ from $\pi_\theta$.

\textbf{Reward design and curriculum learning}: We use multiple levels of curriculum learning to gradually encourage our agent to explore the environment, and a sparse reward function that gives the agent a reward of $+1$ only when it finds the target.

\begin{figure*}[!h]
\centering
\subfigure{\includegraphics[width=2.35in]{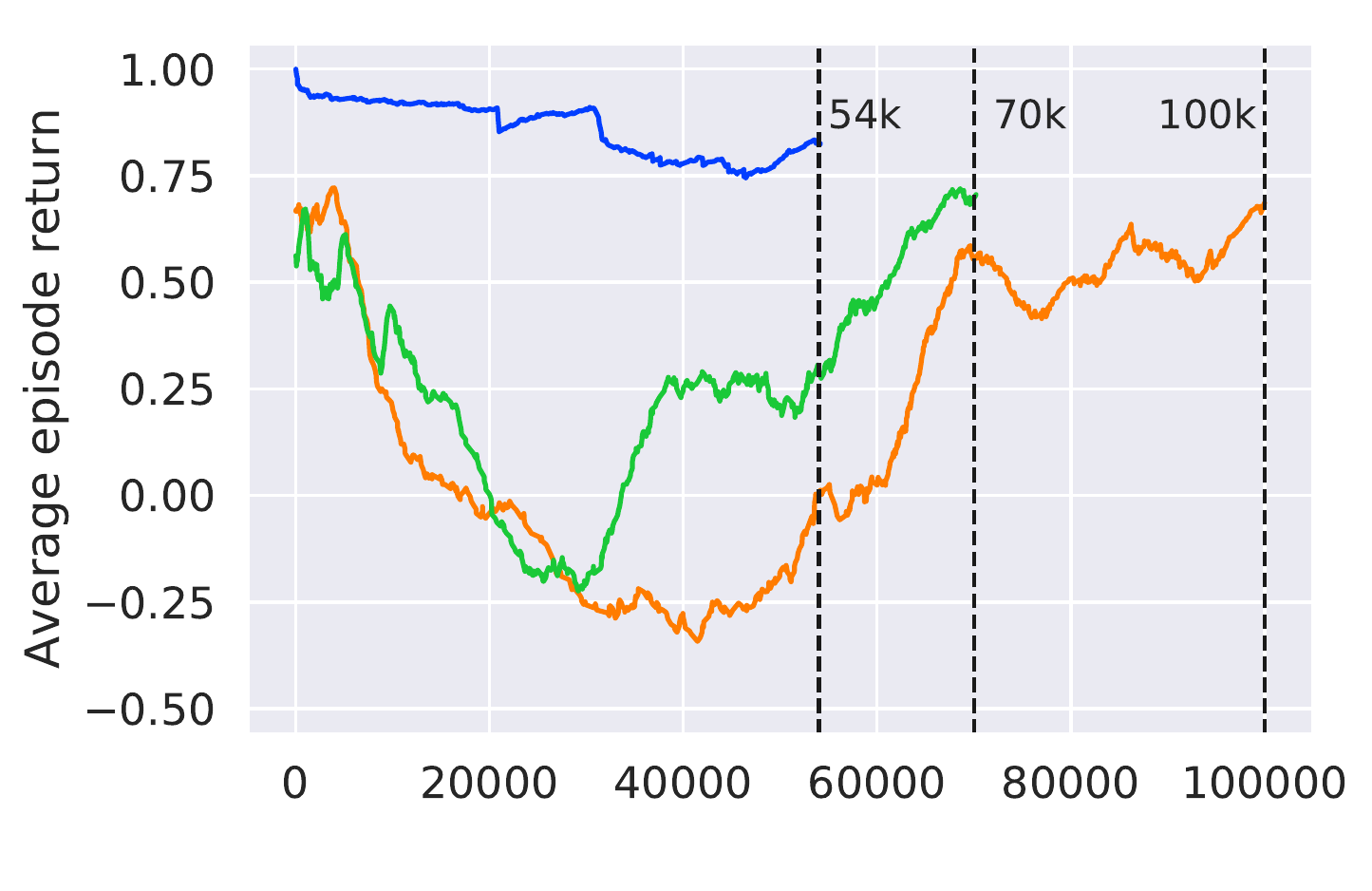}}
\subfigure{\includegraphics[width=2.35in]{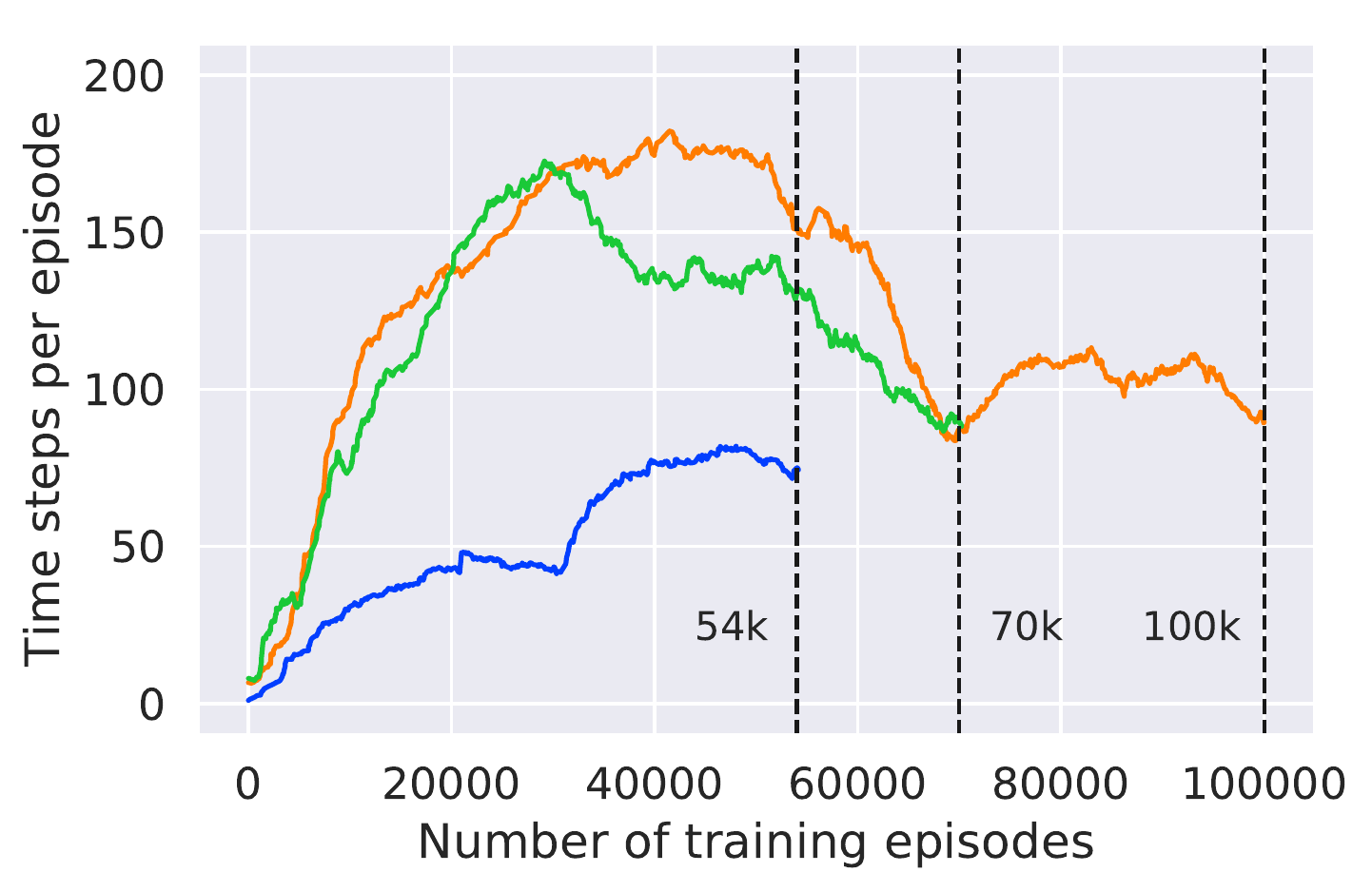}}
\subfigure{\includegraphics[width=2.35in]{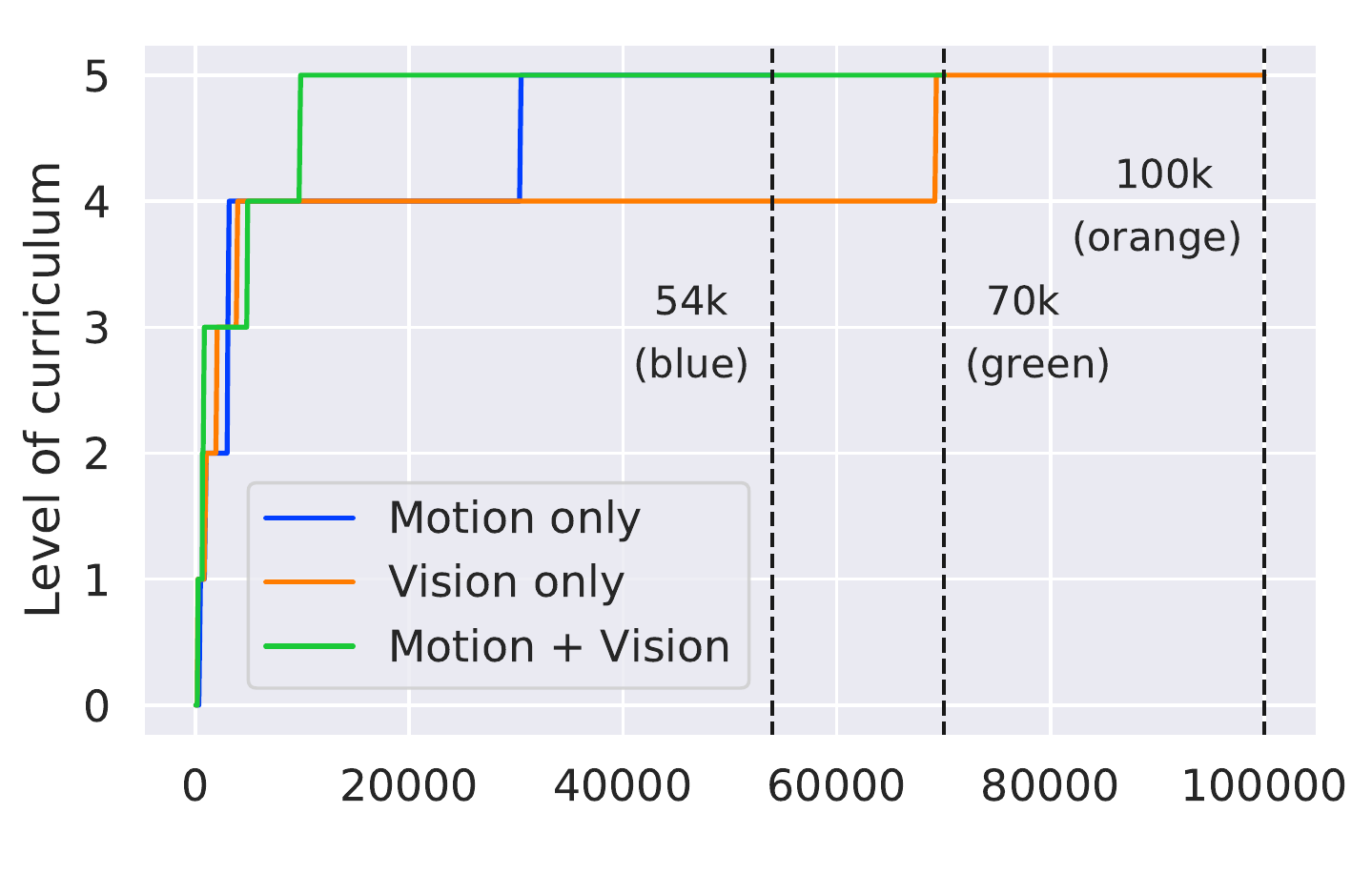}}\\
\vspace{-4mm}
\caption{\textbf{RL training curves on the KITTI dataset}. Our approach incorporates the desirable properties of motion- and vision-only methods for navigation tasks. Using images (alone) seems to increase complexity and reduce performance, but when combining it with motion data we compensate these shortcomings.}
\label{rl-train-kitti}
\vspace{-4mm}
\end{figure*}

\begin{figure*}[!h]
\centering
\subfigure{\includegraphics[width=2.35in]{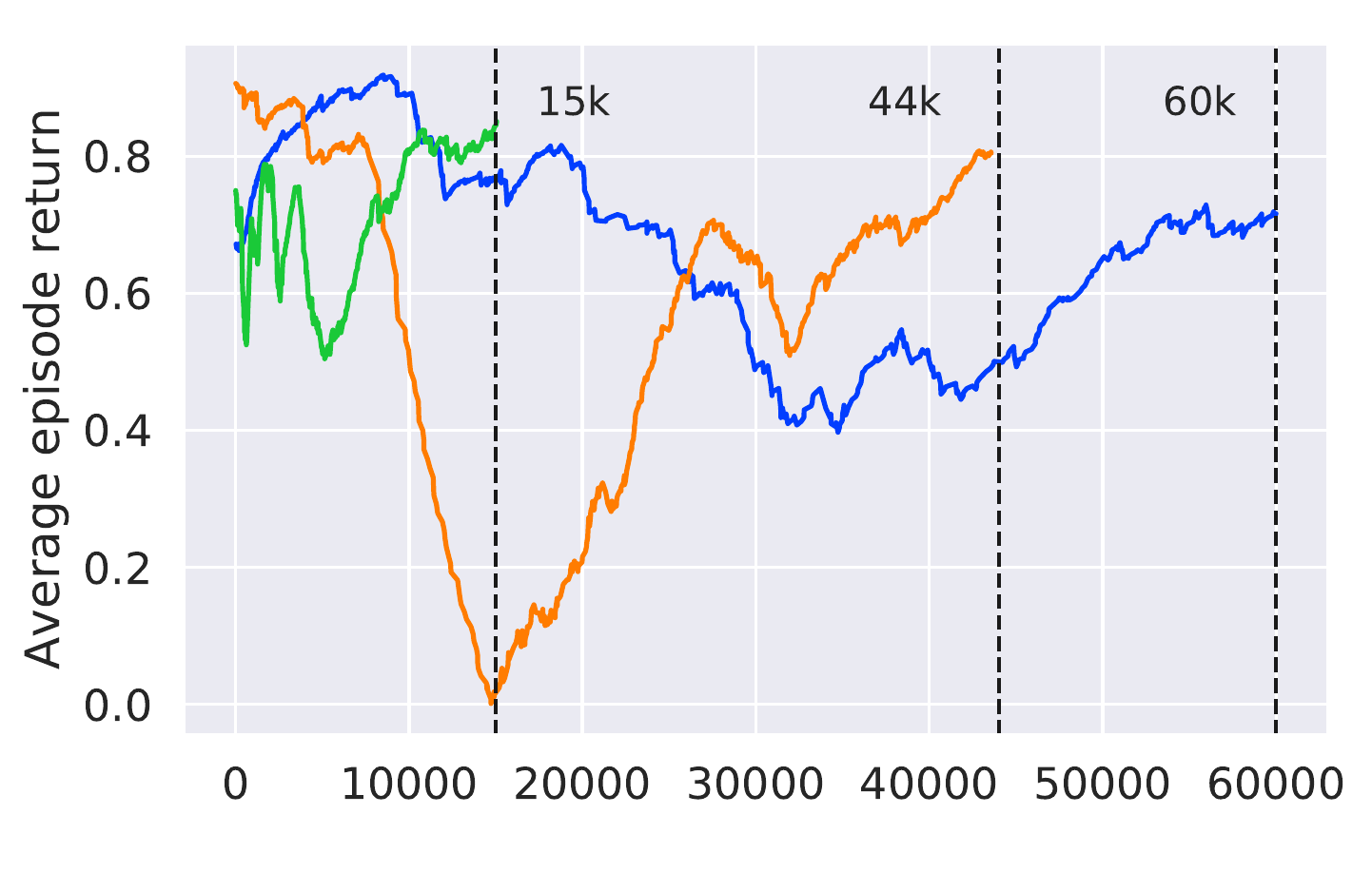}}
\subfigure{\includegraphics[width=2.35in]{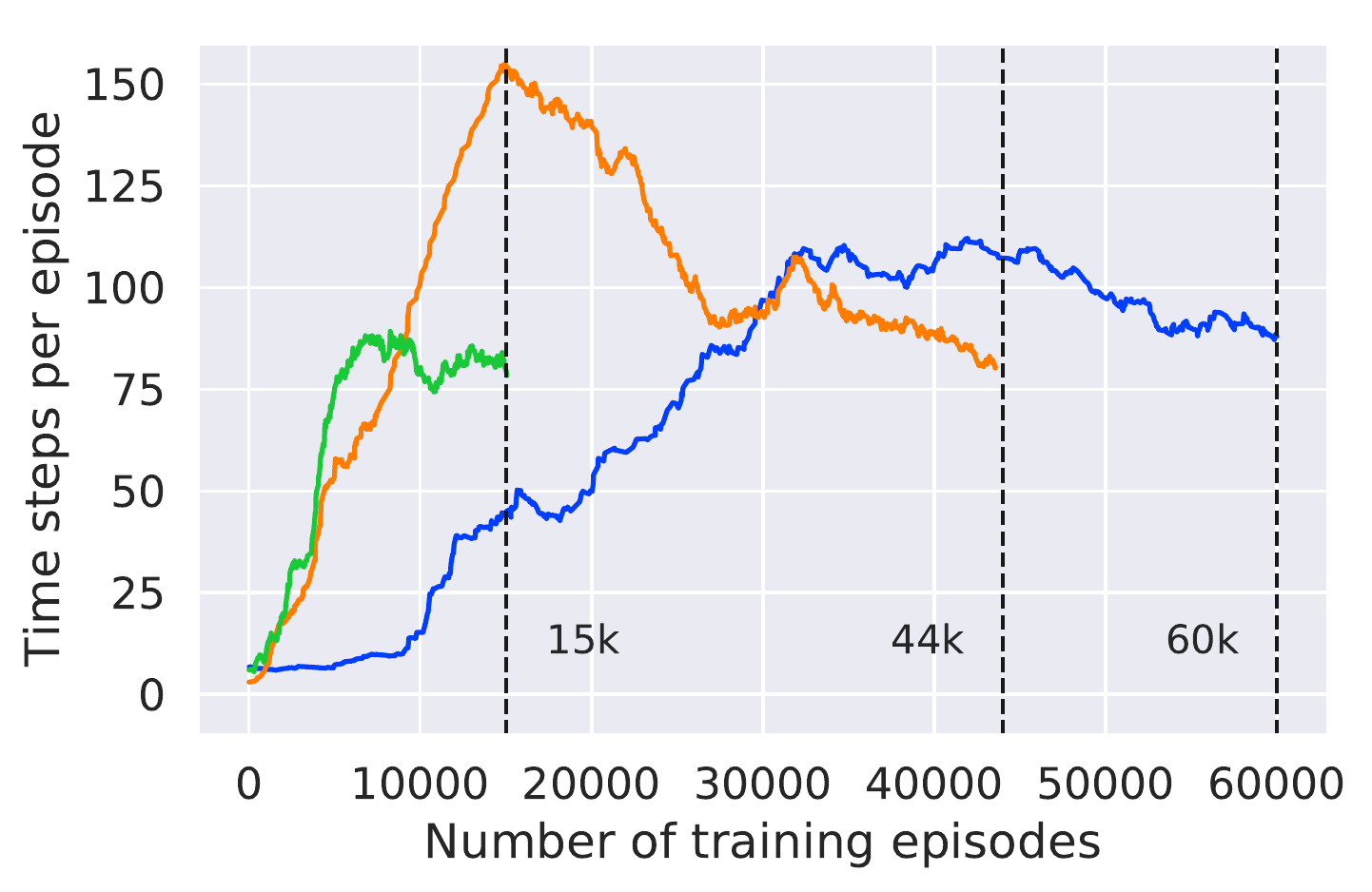}}
\subfigure{\includegraphics[width=2.35in]{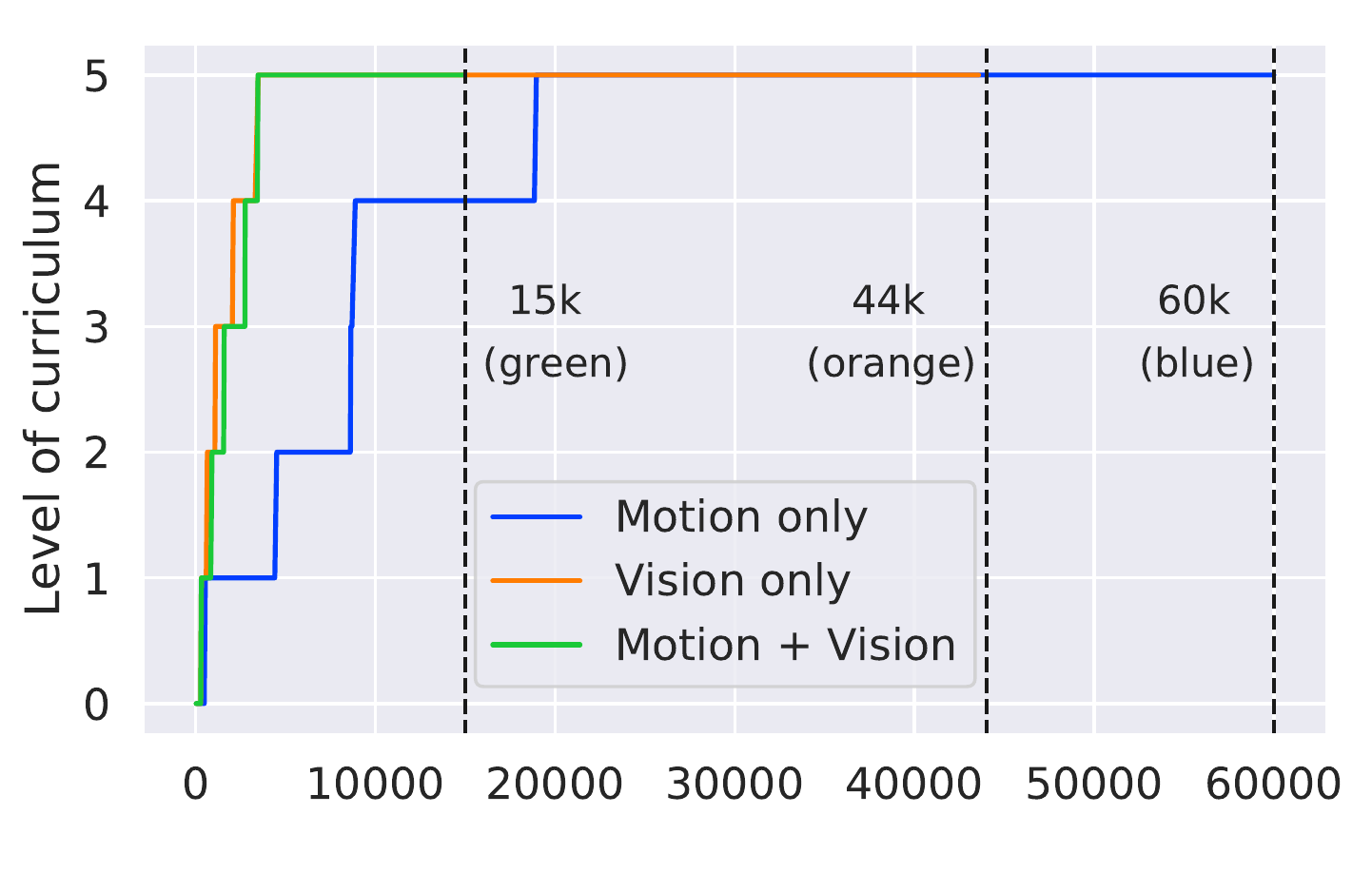}}\\
\vspace{-4mm}
\caption{\textbf{RL training curves on the Oxford RobotCar dataset}. In contrast to the results in Fig. \ref{rl-train-kitti}, we found that using motion+visual data can actually boost the RL training. In this case, our full model required 15k training episodes, compared to 60k and 44k for the motion- and vision-only baselines.}
\label{rl-train-oxf}
\vspace{-2mm}
\end{figure*}

\begin{figure}[!t]
   \centering
   \includegraphics[width=\columnwidth]{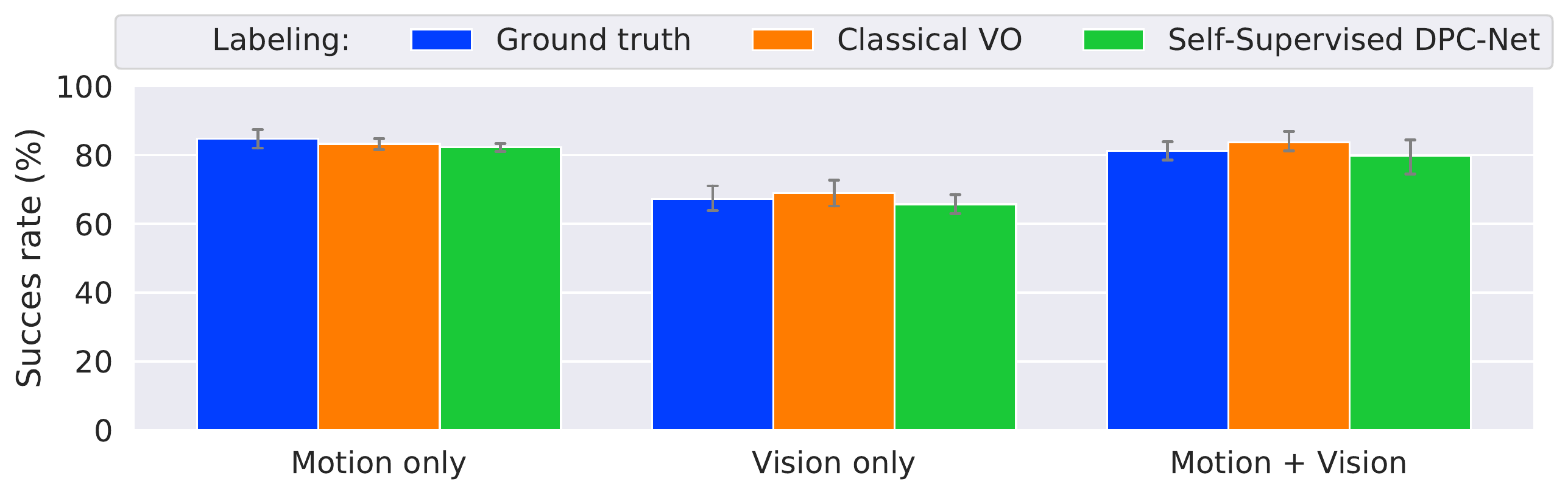}
   \vspace{-6mm}
   \caption{\textbf{RL deployment statistics on the KITTI dataset}. We trained and tested on sequence 05 of the raw data.}
   \label{deploy_kitti_st}
   \vspace{-4mm}
\end{figure}

\section{Experiments}

We evaluate our model on two real driving datasets, Oxford RobotCar \cite{IEEEexample:RobotCarDatasetIJRR} and KITTI \cite{Geiger2013IJRR}, using the CityLearn environment \cite{IEEEexample:chancan2020citylearn}, see Figs. \ref{kitti-deploy} and \ref{ox-deploy}. We additionally conduct experiments to obtain the trade-off between the RL success rate and the motion estimation precision. 

Figs. \ref{rl-train-kitti} and \ref{rl-train-oxf} provide the corresponding quantitative results averaged over 6 runs with different random seeds. We compare our full model (green) with two baselines which correspond to pure motion-only RL (blue), and vision-only RL (orange). These two baselines use the same setup as the full model, except that they only use either motion estimate data or visual observations, respectively, as shown in Fig. \ref{approach}.

\begin{figure}[!t]
   \centering
   \subfigure{\includegraphics[width=\columnwidth]{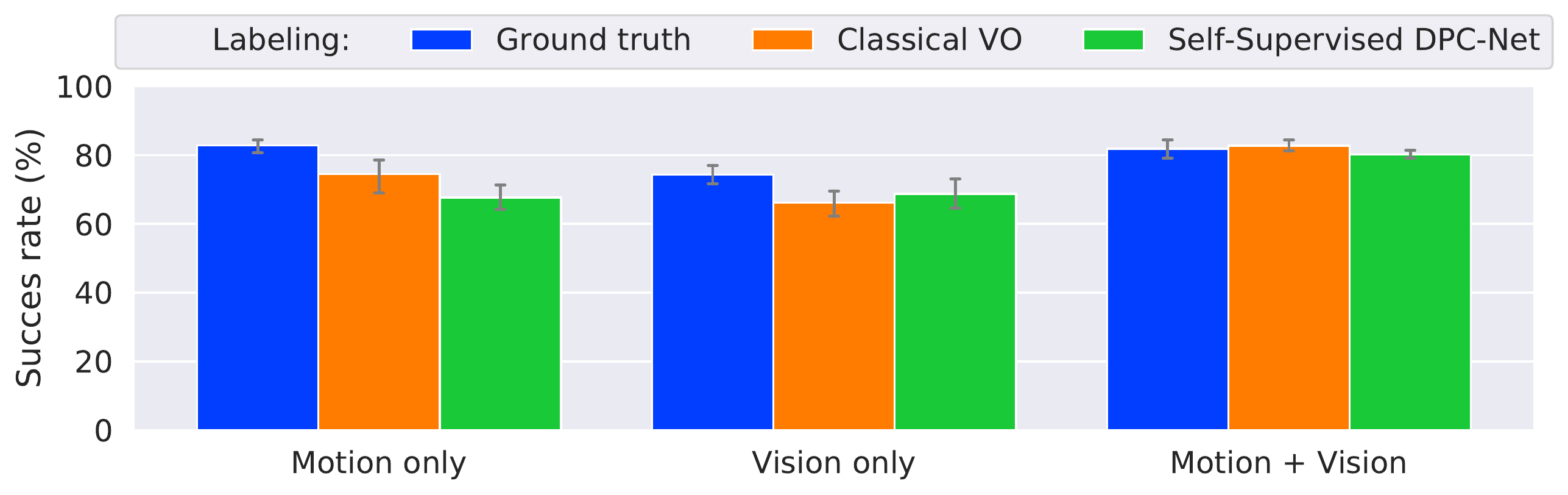}}\vspace{-2mm}
   \centering
   \subfigure{\includegraphics[width=\columnwidth]{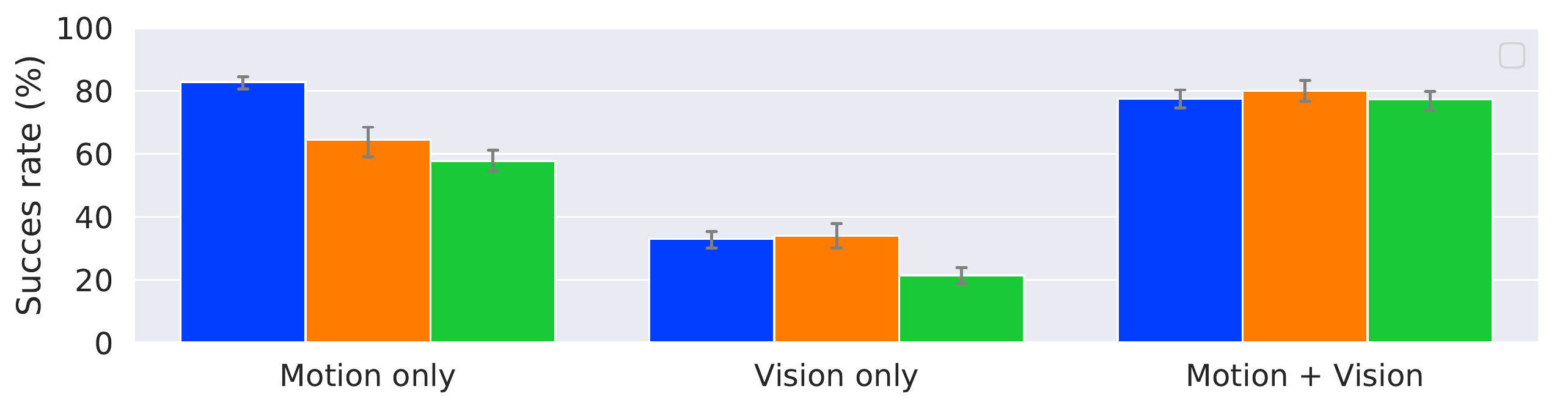}}
   \vspace{-6mm}
   \caption{\textbf{RL deployment statistics on the Oxford RobotCar dataset} on the traversal it was trained (top) and generalizing at night (bottom).}
   \label{deploy_oxf_st}
   \vspace{-4mm}
\end{figure}

We also report deployment statistics in Figs. \ref{deploy_kitti_st} and \ref{deploy_oxf_st}. For the KITTI dataset, our full model can solve navigation tasks with 80\% success rate, compared to 65\% for the vision-only system. In contrast, the agent using motion states seems to compete with our full model, however, its main limitation is that it does not incorporate visual information for distinguishing between environmental changes. For the Oxford RobotCar dataset, where we test generalization from day to night, our full model is capable of consistently obtaining around an 80\% success rate, compared to 30\% for the vision-only system.

\begin{figure}[!t]
   \centering
   \subfigure{\includegraphics[width=0.585\columnwidth]{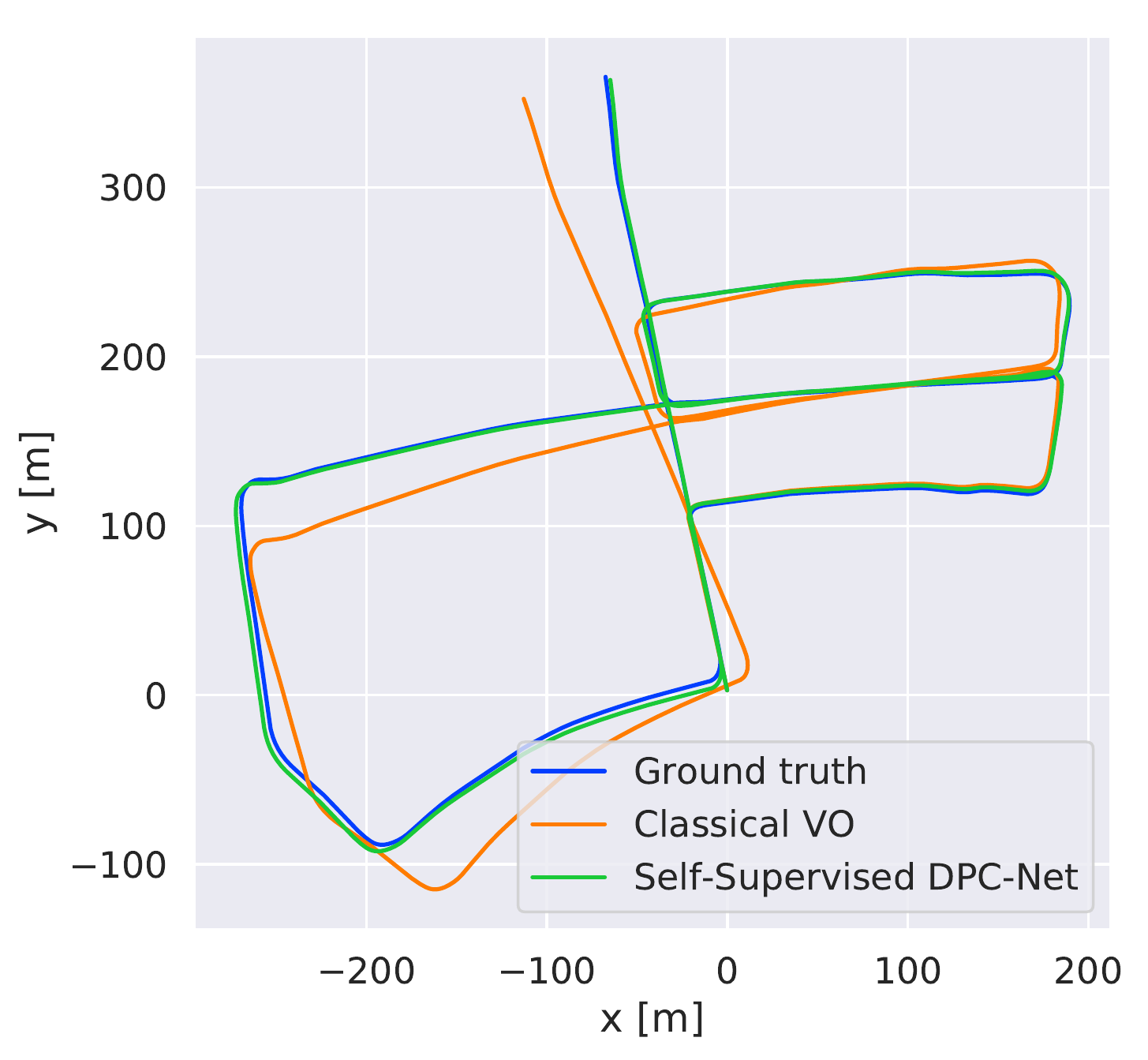}}
   \subfigure{\includegraphics[width=0.375\columnwidth]{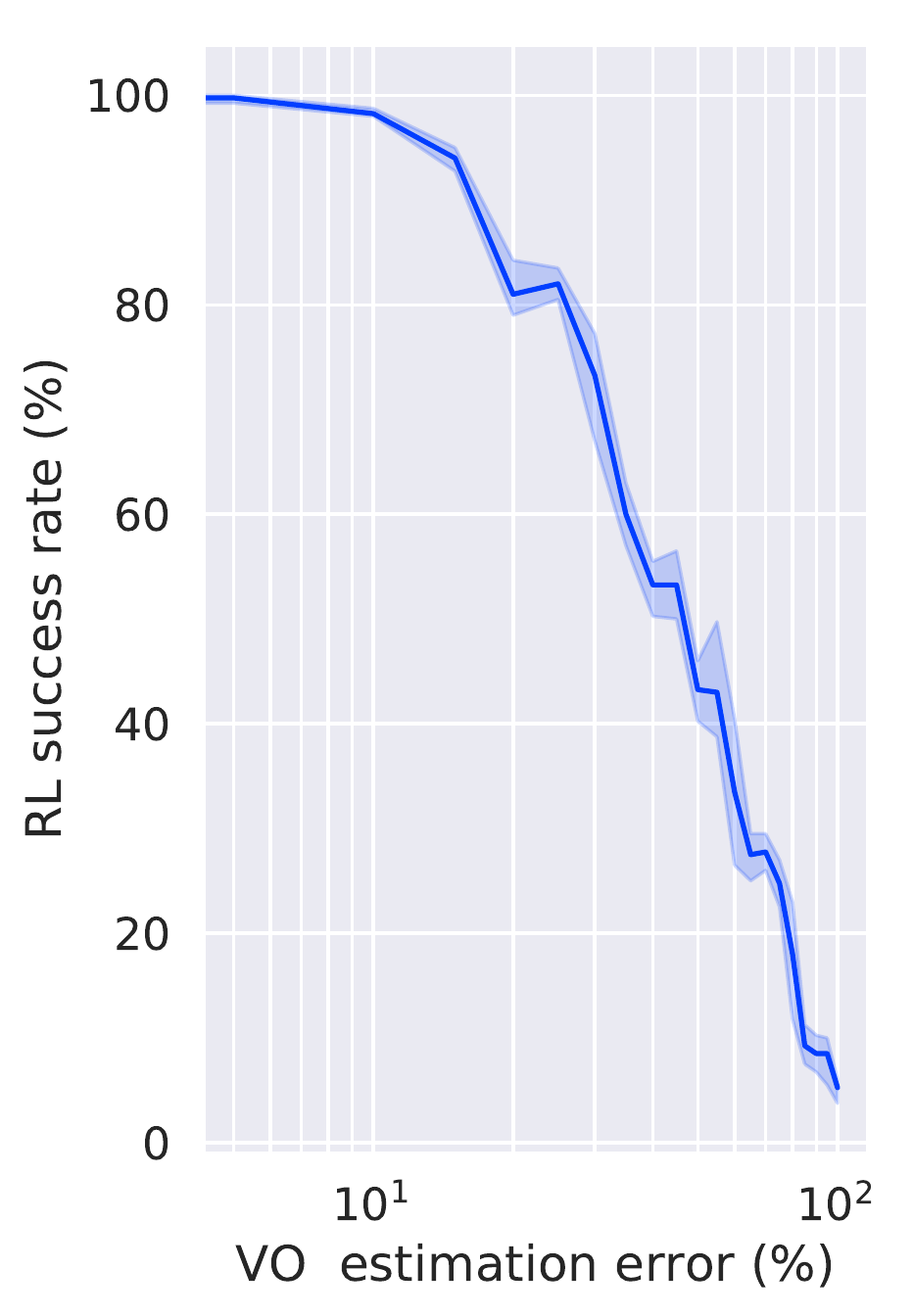}}
   \vspace{-2mm}
   \caption{\textbf{Influence of motion estimation precision}. Ground truth and VO-based data of the KITTI dataset (left). Trade-off between the RL navigation success rate and the ego-motion estimation precision (right).}
   \label{odo_kitti}
   \vspace{-6mm}
\end{figure}

To further analyze the influence of motion estimation precision, in all our experiments we compare the ego-motion data obtained using classical VO and SS-DPC-Net against ground truth data provided by each dataset, see Fig. \ref{odo_kitti} (left) for the KITTI dataset. Interestingly, the difference between these ego-motion results does not seem to impact our three baselines on the KITTI dataset, as all these models are deployed on the same traversal used for training. Conversely, on the Oxford RobotCar dataset, as we also deploy under drastic visual changes (day to night), we note that our full model retains good navigation performance, compared to vision-only systems. We also provide insights on the influence of the VO precision to our full model in Fig. \ref{odo_kitti} (right).

\section{Conclusion} 
\label{sec:conclusion}

We have shown that combining self-supervised learning for visuomotor perception and RL for decision-making considerably improves the ability to deploy robotic systems capable of solving complex navigation tasks from raw image sequences only. We proposed a method, including a new neural network architecture, that temporally integrates two fundamental sensor modalities such as motion and vision for large-scale target-driven navigation tasks using real data via RL. Our approach was demonstrated to be robust to drastic visual changing conditions, where typical vision-only navigation pipelines fail. This suggest that odometry-based data can be used to improve the overall performance and robustness of conventional vision-based systems for learning complex navigation tasks. In future work, we seek to extend this approach by using unsupervised learning for both decision-making and perception.



\bibliographystyle{plainnat}
\bibliography{references}

\end{document}